ARTICLE

# Ranking Facts for Explaining Answers to Elementary Science Questions

Jennifer D'Souza[1], Isaiah Onando Mulang'[2], Sören Auer[1]

[1]TIB Leibniz Information Centre for Science and Technology, Hannover, Germany
[2]University of Bonn, Bonn, Germany

{jennifer.dsouza | auer}@tib.eu
mulang.onando@ibm.com

## Abstract

In multiple-choice exams, students select one answer from among typically four choices and can explain why they made that particular choice. Students are good at understanding natural language questions and based on their domain knowledge can easily infer the question's answer by 'connecting the dots' across various pertinent facts.

Considering automated reasoning for elementary science question answering (Clark et al. 2018), we address the novel task of generating explanations for answers from human-authored facts (Jansen and Ustalov 2019). For this, we examine the practically scalable framework of feature-rich support vector machines leveraging domain-targeted, hand-crafted features. Explanations are created from a human-annotated set of nearly 5,000 candidate facts in the WorldTree corpus (Jansen et al. 2018). Our aim is to obtain better matches for valid facts of an explanation for the correct answer of a question over the available fact candidates. To this end, our features offer a comprehensive linguistic and semantic *unification paradigm*. The machine learning problem is the preference ordering of facts, for which we test pointwise regression versus pairwise learning-to-rank.

Our contributions, originating from comprehensive evaluations against nine existing systems, are: (1) a case study in which two preference ordering approaches are systematically compared, and where the pointwise approach is shown to outperform the pairwise approach, thus adding to the existing survey of observations (Kamishima et al. 2010; Melnikov et al. 2016) on this topic; (2) since our system outperforms a highly-effective TF-IDF-based IR technique (Chia et al. 2019) by 3.5 and 4.9 points on the development and test sets, respectively, it demonstrates some of the further task improvement possibilities (e.g., in terms of an efficient learning algorithm, semantic features) on this task; (3) it is a practically competent approach that can outperform some variants of BERT-based reranking models (Banerjee 2019; Chia et al. 2019); and (4) the human-engineered features make it an interpretable machine learning model for the task.

## 1. Introduction

There is an emerging trend in AI surrounding explanation generation which aims to improve the *interpretability* of the machine learning process (i.e., to reveal how the system arrives at the prediction) (Lundberg and Lee 2017; Paul and Frank 2019), or *trustworthiness* of the result (i.e., to make the prediction more believable as correct by justifying it) (Ribeiro et al. 2016; Jansen et al. 2018). On the one hand, interpretability enables better comprehension of the so-called black box machine learning, while on the other hand, trustworthiness determines usefulness of the technology in decision critical domains such as in medicine and law. In other words, machine learning predictions cannot be acted upon in such domains without additional supporting evidence, as the consequences may be dramatic.

Concerning interpretability, it is debatable whether it is possible in reality to generate complete descriptions of complex systems as explanations, where the explanations would most likely only describe a simplified version of the actual system (Mittelstadt et al. 2019). In a sense, interpretability subsumes trustworthiness, but not completely, since trustworthiness can be concerned to a lesser extend with why the model made the decision but rather with why the decision is *true* using additional world and commonsense knowledge (Bauer et al. 2018). This view of enabling better trustworthiness of a machine learned result via explanations is the focus of this work, which we apply to the domain of elementary science question answering (QA) over standardized tests. Our task is: given a corpus of elementary science question and correct answer pairs ('QA pairs' hence) taken from standardized tests, to automatically justify the correct answer with

---

**Question** A student put 200 milliliters (mL) of water into a pot, sets the pot on a burner, and heats the water to boil. When the pot is taken off the burner, it contains only 180 milliliters (mL) of water. What happened to the rest of the water?

**Answer** it turned into water vapor

**Explanation**

(f1) to turn means to change

(f2) water is in the gas state, called water vapor, for temperatures between 373 or 212 or 100 and 100000000000 k or f or c

(f3) boiling or evaporation means change from a liquid into a gas by adding heat energy

(f4) water is a kind of liquid

(f5) evaporation causes amount of water to decrease

(f6) a burner is made of metal

(f7) a burner is a part of a stove

(f8) a stove generates heat for cooking usually

(f9) pot or pan or frying pan is made of metal for cooking

(f10) metal is a thermal or thermal energy conductor

(f11) a thermal energy conductor transfers heat from warmer objects or hotter objects to cooler objects

(f12) if a thermal conductor or an object is exposed to a source of heat then that conductor or that object may become hot or warm

(f13) a source of something emits or produces or generates that something

(f14) if one surface or one substance or one object touches something then one is exposed to that something

(f15) being on something or placed in something or placed over something means touching that something

(f16) heat energy is synonymous with thermal energy

(f17) transferring is similar to adding

(f18) conductivity is a property of a material or substance

(f19) if an object is made of a material then that object has the properties of that material

(f20) metal is a kind of material

(f21) a burner is a kind of object or surface

---

Table 1. : Example Instance in the WorldTree Corpus (Jansen et al. 2018). A Question and Correct Answer pair (QA pair) with its Explanation comprising 21 logically ordered facts (f1, f2, ..., f21). In the WorldTree, explanation lengths vary between 1 and 21 facts. This selected example Explanation with 21 facts is the longest in the corpus. Characteristic of the data design, facts in explanations lexically overlap (shown as underlined words) with the question or answer or other facts.

an explanation generated from science and commonsense facts. An example of a QA pair and its explanation is illustrated in Table 1. We generate these explanations from facts taken from the WorldTree Corpus (Jansen et al. 2018) – a newly released, manually authored knowledge base of semi-structured tables (also called 'a tablestore') containing nearly 5,000 elementary science and commonsense facts.

The introduction of the WorldTree corpus (2018) presents a new direction for evaluating machine intelligence. In the task defined by the corpus, systems can be evaluated w.r.t. their language understanding, reasoning, and use of common-sense knowledge capacities via the generated explanations. These are new opportunities to advance the state-of-the-art in machine intelligence w.r.t. natural language understanding in a similar vein to the Turing test (Turing 2009).

Testing machine intelligence in natural language inference tasks over standardized tests was first initiated by the AI2 ARC challenge (2018), that originally released just the QA part of the corpus. This challenge has helped to move forward the reasoning abilities of natural language inference systems on tasks that children can accomplish, while (ideally) increasing their ability to explain their reasoning. Progress largely stalled in the fifty percent accuracy for years with some notable exceptions (Parikh et al. 2016; Seo et al. 2016; Khot et al. 2018), and then large language models were shown to reach >90% performance (Clark et al. 2019), but did so without producing explanations. The WorldTree corpus (2018) fills the gap by providing a way of explicitly measuring the explanation-generation ability of a model on the ARC corpus, as well as training the models to perform the many-fact explanation-generation task.

Generally, in multiple-choice QA exams, a student selects one answer to each question from among typically four choices and can explain why they made that particular choice based on their world and commonsense knowledge. For a machine, on the other hand, constructing an explanation for the correct answer can be challenging for the following reasons: 1) It can be a multi-step process since some facts may directly relate to the question and correct answer, but there may be others that build on the earlier facts provided as explanation. Consider in Table 2, facts f1 and f2 directly relate to the question and correct answer; however, fact f3 is an elaboration for f2. This phenomenon is even more prevalent in longer explanations. Consider the example in Table 1, where facts f6 to f14 are indirectly related to the question or correct answer, nonetheless are essential to the logical sequence of facts to explain the phenomenon of "heating of water caused by the pot on the burner." And, 2) this multi-step inference is highly amenable to the phenomena of *semantic drift*, i.e. the tendency of composing spurious inference chains leading to wrong conclusions (Khashabi et al. 2019; Fried et al. 2015). This is depicted by the facts in red in Table 2, that on the surface are linguistically related to the question and correct answer, but are not semantically relevant to the explanation for the correct answer.

**Question** Granite is a hard material and forms from cooling magma. Granite is a type of
**Answer** igneous rock
**Explanation**
(f1) igneous rocks or minerals are formed from magma or lava cooling;
(f2) igneous is a kind of rock;
(f3) a type is synonymous with a kind;
rock is hard;
to cause the formation of means to form;
metamorphic rock is a kind of rock;
cooling or colder means removing or reducing or decreasing heat or temperature;

Table 2. : Example depicting lexical hop between Question and Correct Answer pair not just with correct facts, but also with incorrect fact candidates.

In this work, we address the aforementioned machine learning challenges by simultaneously expanding both the linguistic and conceptual vocabulary of the question, correct answer, and explanation fact words, in a domain-targeted manner as features for machine learning. By expanding the vocabulary, we aimed to obtain greater number of lexical matches between the QA pair and explanation facts. In this way, we also indirectly aimed to facilitate improved semantic relatedness between the QA pair and their explanation facts via their expected greater number of lexical matches. Overall, six differing and novel information categories were leveraged to represent the instances for learning. While in an earlier system (D'Souza et al. 2019), we have similarly employed a feature-based approach for this task, in our new version presented in this article, the generic features of that system are replaced by a domain-targeted set.

With respect to the machine learning strategy, we adopt the *learning how to order* problem formulation since the annotated explanations in the WorldTree corpus (2018) are made up of logically ordered facts in discourse. Specifically, in the context of the WorldTree, the automatic task entails learning and predicting preferences over candidate facts per QA pair explanation. Generally, learning a preference function involves ranking facts from a candidate set, i.e. the relevant facts before the irrelevant facts, and the relevant facts in order w.r.t. each other. Further, it also includes an implicit "abstaining" from making ranking decisions between the irrelevant facts. Then during testing, new QA pair explanations are generated by predicting the order for the facts using the trained preference function. Since the problem does not involve a total ordering of all facts in the tablestore for the explanations, but only the relevant facts, we adopt the preference learning approach (Fürnkranz and Hüllermeier 2010; Kamishima et al. 2010) rather than a ranking approach, where the latter entails a total ordering. Nevertheless, preference ranking are a class of problems that subsume ranking functions. In fact, among the problems in the realm of preference learning, the task of "learning to rank" has probably received the most attention in the literature so far, and a number of different ranking problems have already been introduced. In this work, we compare a pointwise preference learning approach versus the pairwise ranking approach. Further, the scoring and loss functions for both pointwise and pairwise ranking is from the support vector machine class of learning algorithms. Support vector machines are preferred by many as a strong classifier needing less computation power than neural models. Although we are not the first to contrast pointwise and pairwise learning, our study offers new observations on the comparison of these two techniques on a new problem, i.e. the ranking of facts to construct explanations. In this way, we build on our earlier system (2019) that tested only the pairwise ranking approach with its generic features set.

We conduct extensive empirical evaluations of our proposed approach with nine existing systems. Our main contributions are:

- Insights into the comparison between a pointwise and a pairwise machine learning technique for constructing explanations as a preference learning problem, thus presenting a new observation complementing existing studies (Melnikov et al. 2016; Kamishima et al. 2005 2010) in this field;
- A domain-targeted space of representative features of world and commonsense knowledge to associate a QA pair and candidate explanation facts both linguistically and semantically. Consequently, our feature-based model is human-interpretable. Further, empirical evaluations show that our model effectively outperforms standardized BERT-based (Devlin et al. 2018) neural techniques, that in contrast to ours, are seen as uninterpretable black-box models.

The rest of the article is structured as follows. We first describe the corpus we use (Section 2) and the task (Section 3). After that, we describe the related work (Section 4). Section 5 details our approach followed by a discussion on the features in our feature-rich approach in Section 6. Our experimental results and analysis are then presented (Section 7). Finally, we conclude with future directions in Section 8.

## 2. Corpus

The data used in this study comes from the WorldTree corpus[a] (Jansen et al. 2018). It comprises a portion of the standardized elementary science exam questions, 3rd to 5th grades, drawn from the Aristo Reasoning Challenge (ARC) corpus (Clark et al. 2018). The questions have multiple choice answers with the correct answer known. Each question-correct answer pair (QA pair) in the WorldTree corpus (2018) has detailed human-annotated explanations, consisting of between 1 to 21 facts that are arranged in logical discourse order w.r.t. each other. The QA pair instances are divided in the standard ARC train, development, and test sets. The WorldTree corpus then is provided as 1,190 training, 264 development, and 1,248 test instances where each instance is a QA pair and its explanation. In all, 14.4% of the training fold facts, 40.4% of the development fold facts, and 22.5% of test fold facts are overlapping.

### *2.1 Explanations for Correct Answers to Elementary Science Questions*

As alluded to above, QA pairs in the WorldTree corpus (2018) are annotated with explanations of up to 21 facts (see in Fig. 1 the distribution of facts in the explanations in the training and development sets).

| Total unique explanation facts: | 4,789 |
| Seen in training data: | 2,694 |
| Seen in development data: | 964 |
| Seen in training and development data: | 589 |

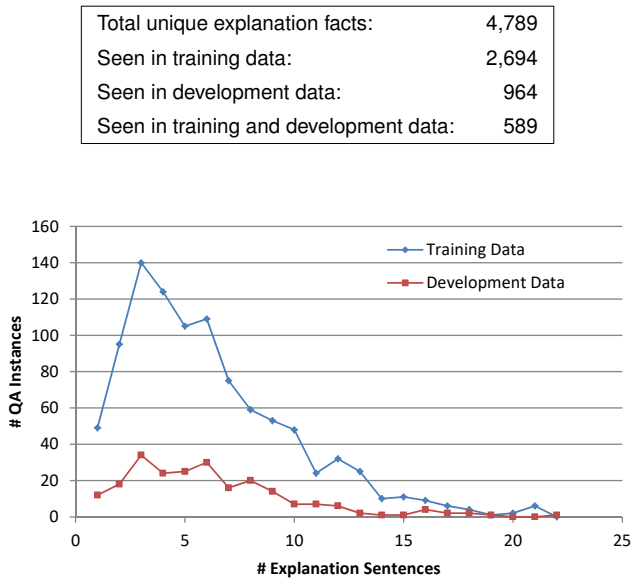

Figure 1: Facts in explanations per question-answer pair in the training and development datasets.

Based on corpus design decisions, the inclusion criteria for facts in explanations were: *lexical overlap*—facts lexically overlap with the question or answer, or with other facts in the explanation; and *coherency*—the explanation facts form a logically coherent discourse fragment. As a consequence of the *lexical overlap* characteristic, a traversal path can be traced between each QA pair and its explanation facts via multiple lexical hops (depicted in Tables 1 and 2 via the underlined words). Further, as an additional annotation layer, facts in each training and development set explanation were categorized as one of three classes. These classes were determined by the role played by the fact in the explanation. Specifically, the classes were *Central*, *Grounding* and *Lexical Glue*. Central facts were defined as core scientific facts relevant to answering the question. E.g., facts such as "as the amount of rainfall increases in an area, the amount of available water in that area will increase." Grounding facts were those which connected to other core scientific facts present in the explanation. E.g., "rain is a kind of water" would connect "rain" and "water" present across two or more Central facts in the explanation. Finally, lexical glue facts expressed synonymy or definitional relationships. E.g., "rainfall is the amount of rain an area receives." Table 3 offers statistics on the overall prevalence of explanation facts across QA pairs in the training and development sets, and also per explanation fact category.

---

[a]We use the TextGraphs2019 Explanation Reconstruction Shared Task Data Release available at http://cognitiveai.org/explanationbank/

| | Total QA pairs | 1,213 |
|---|---|---|
| | Total facts used | 7,448 |
| | Facts per QA pair | 6.14 |
| *Central* | Total facts used | 3,705 |
| | Facts per QA pair | 3.05 |
| *Grounding* | Total facts used | 2,131 |
| | Facts per QA pair | 1.76 |
| *Lexical Glue* | Total facts used | 1,612 |
| | Facts per QA pair | 1.32 |

Table 3. : Corpus statistics for QA pairs w.r.t. their explanation facts from the WorldTree (2018) training and development corpora combined

We now elaborate on the facts' tablestore that formed the reference set for constructing the explanations per QA pair. The tablestore facts were authored based on the elementary science themes of the ARC question-answering data. They are organized in 65 tables representing relation predicates such as *kind of* (e.g., an acorn is a kind of seed), *part of* (e.g., bark is a part of a tree), *cause* (e.g., drought may cause wildfires); or the *actions* of organisms (e.g., some adult animals lay eggs); or the *properties of things* (e.g., an acid is acidic); or *if-then* conditions (e.g., when an animal sheds its fur, its fur becomes less dense). In Table 4, we depict the table types whose facts belonged to at least 1% of the explanations in the training and development sets. We see that only 21 tables from 65 in total were represented in at least 1% of the training and development explanations. Of the 44 remaining least frequently selected tables, in Table 5, we show only nine selected ones as examples. Based on the table sizes, the least frequently occurring tables have fewer facts than most of the 21 frequently selected tables; however, there are one or two exceptions (e.g, the COUNTRY-HEMISPHERE table with 269 facts).

Figure 2 depicts the top six of the 21 frequently selected table types for their fact rankings in the explanations. The six tables are: KINDOF, SYNONYMY, ACTION, IF_THEN, CAUSE, and USED-FOR. For these tables, we explicitly show the proportions of their facts at ranks 1 to 10 in the explanations, and aggregate the remaining lower-ranked facts in a single proportion. In the figure, we can see that except for SYNONYMY, all the remaining tables have a major proportion of their selected facts appear between ranks 1 to 5. For SYNONYMY, however, we see that only 5% of its facts appear at rank 1. This comparatively low proportion is meaningful in the context of the role played by its facts in the explanations - more often than not, they supplement the information of a previous fact. E.g., an explanation has the fact "the moon orbiting the Earth approximately occurs 13 times per year" at rank 1 which is supplemented by the SYNONYMY fact "approximately means about" at rank 2.

| | | | | | |
|---|---|---|---|---|---|
| KINDOF (1120) | 25.22 | REQUIRES (122) | 2.87 | ATTRIBUTE-VALUE-RANGE (70) | 1.53 |
| SYNONYMY (636) | 14.27 | PARTOF (149) | 2.74 | | |
| ACTION (260) | 6.48 | COUPLEDRELATIONSHIP (127) | 2.67 | CHANGE (63) | 1.53 |
| IF-THEN (230) | 5.31 | | | CHANGE-VEC (63) | 1.43 |
| CAUSE (184) | 4.17 | SOURCEOF (82) | 1.89 | EXAMPLES (59) | 1.43 |
| USEDFOR (192) | 4.17 | CONTAINS (76) | 1.79 | PROPERTIES-GENERIC (47) | 1.21 |
| PROPERTIES-THINGS (174) | 3.58 | AFFECT (78) | 1.73 | TRANSFER (47) | 1.11 |
| | | MADEOF (73) | 1.69 | AFFORDANCES (49) | 1.08 |

Table 4. : Percentage occurrences of facts from 21 table types (of 65 total) that participated in at least 1% of the training and development explanations. The numbers in parenthesis are the table sizes.

| | | | | | |
|---|---|---|---|---|---|
| OPPOSITES (36) | 0.978 | DURING (31) | 0.684 | PREDATOR-PREY (3) | 0.065 |
| LOCATIONS (48) | 0.912 | MEASUREMENTS (24) | 0.587 | COUNTRY-HEMISPHERE (269) | 0.065 |
| FORMEDBY (41) | 0.782 | CONVERSIONS (4) | 0.098 | PERCEPTIONS (13) | 0.033 |

Table 5. : Nine randomly chosen tables as examples with less than 1% percentage fact occurrences (of 44 in all) in the training and development explanations. The numbers in parenthesis are the table sizes.

## 3. The Explanation Regeneration Task Description

Our task is defined after the TextGraph-19 Shared Task on Explanation Regeneration (Jansen and Ustalov 2019) where the WorldTree corpus (2018) was leveraged for the first time to facilitate machine learning system development. It was posited as an ordering task as follows.

For a QA pair, given an unordered collection of facts (in our case, the 4,789 tablestore of facts), the task objective is to order the given collection (as shown in Tables 1 and 2), such that the relevant facts will be top-ranked w.r.t. the irrelevant facts; and, further, the top-ranked relevant facts will be in a logical discourse order w.r.t. each other. Note that the irrelevant facts will also be returned; however, for the task it is sufficient that they are lower-ranked than the relevant ones. Formally, given a question '$q$,' its known correct answer '$a$,' and an unordered collection of facts $F_{uno}$, the ordering objective is to (1) determine all facts $\in F_{uno}$ that are relevant to the $(q, a)$ pair, and (2) order the relevant facts to form a logically ordered discourse fragment thus producing a partially rank ordered collection $F_{po}$. The resulting collection is seen as partially rank-ordered since only the ordering for the relevant '$k$' facts to the QA pair is a meaningful result; the remaining '$|F_{po}|$-k' facts that are consequently ordered as a result of applying the learned function to the full facts' tablestore remain irrelevant to the given QA pair. Thus, at a high level, the task can be viewed as explanation regeneration since each QA pair initially gets the entire collection of facts as an explanation which it then must order by preference for relevance and discourse.

## 4. Background and Related Work

***Reasoning in Elementary Science QA entailing various knowledge sources***. Clark et al. (2013) and Jansen et al. (2016) in their respective studies found three main question categories in Elementary Science QA: 1) retrieval questions relying on taxonomic, definitional, or property knowledge; 2) inference questions tapping into knowledge of causality, processes, or specific instances of occurrences; and 3) domain-specific questions.

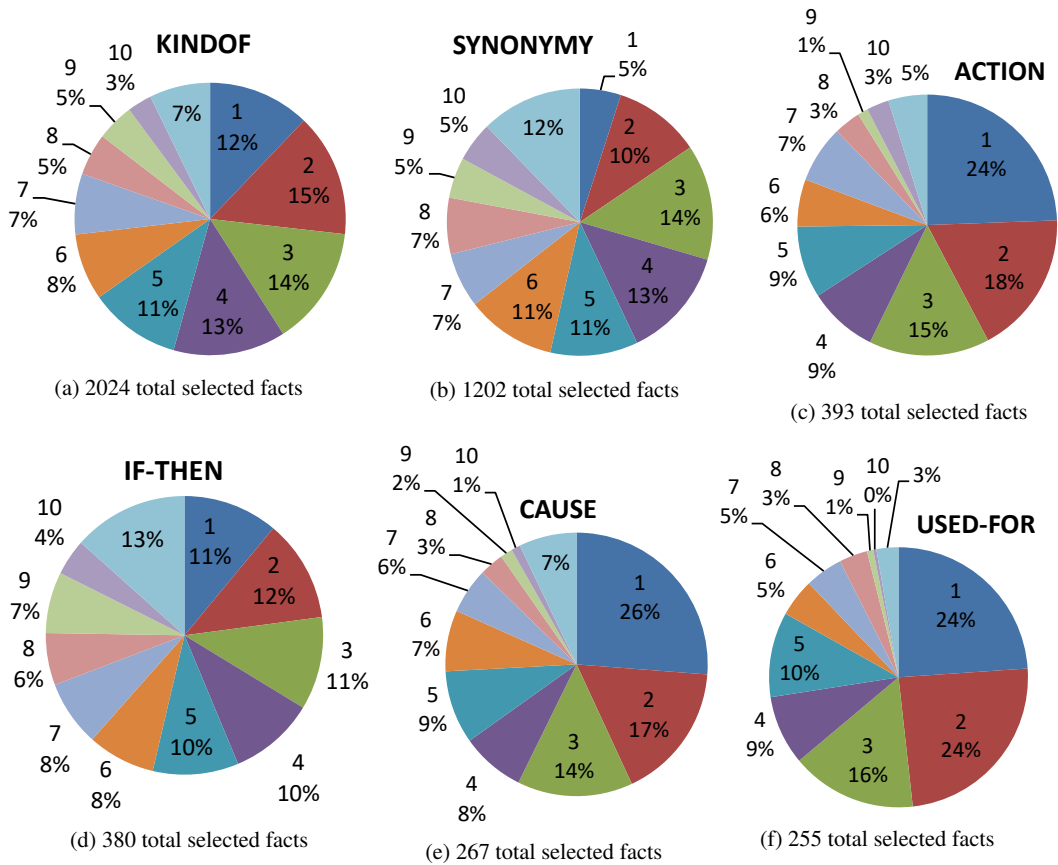

Figure 2: Ranked proportions of a subset of facts from the training and development set explanations. These facts are only those in the top-6 most frequently selected fact table types (shown as Figures 2a to 2f, respectively). Each pie-chart depicts the respective % proportion of occurrences of facts for ranks 1 to 10 explicitly, with one additional unlabeled category for the ranks after 10.

Thus to cater to the different question categories, reasoning in Elementary Science QA entails reliance on various knowledge sources extending the linguistic information of the QA strings themselves. To this end, our proposed system employs *lexical*, *grammatical*, and *semantic* feature categories that, respectively, facilitate addressing the QA types found in the aforementioned studies. For instance, we employ *lexical* features that expand the given word vocabulary and *grammatical* features as an abstraction of the role of words in a sentence. The expanded vocabulary facilitates matches between synonyms and word types which we observed as a characteristic of many of the retrieval questions in our corpus – they do not match directly with a relevant explanation fact but indirectly via synonyms. Our *semantic* feature category including commonsense and knowledge embeddings ensures semantic coherence, beyond simple lexical matching between explanation sentences addressing taxonomic, definitional, or property knowledge, relevant also in inference or domain-specific questions.

***Commonsense Knowledge for Explanations***.    ConceptNet (Speer et al. 2017) as a commonsense knowledge source was employed in two recent systems (Paul and Frank 2019; Bauer et al. 2018) addressing inference tasks involving either explanation construction or narrative generation. Paul and Frank's (2019) system traced multi-hop paths across ConceptNet's entities as features to boost performance of predicting human need categories, and in turn the paths when traced seemed a reasonable explanation for the result in most cases Bauer et al. (2018). For a narrative QA task, employed multi-step reasoning by tracing a path via ConceptNet between entities in the question and entities in a given context where the selected contextual sentences were then posited as the question's answer in narrative form.

Relatedly, our data domain, i.e. Elementary Science QA, poses questions on entities like 'sun,' 'fire,' 'friction,' etc.—themes of varying degrees of abstraction that are often non-named entities. By leveraging ConceptNet, we were interested in obtaining additional qualifying information about these terms thereby to better match a QA with its related explanation facts. E.g., a 'sun' is a type of a 'star.' Unlike earlier works (2018; 2019) that gathered information from paths traced through the ConceptNet graph, for our data domain and purpose, we queried just for additional qualifying commonsense information about the terms in the QA and the explanation facts themselves, which were then checked for matches. These qualifying features, in turn, served as the conceptual glue between disparate units of information. Additionally, a path traced through the ConceptNet graph between QA and explanation fact terms could have been a viable feature, but we have not tried it as the term pair combinations between the QA and the explanation facts could have made the computation of our features forbiddingly complex.

*Generating Explanations for Elementary Science QA*.   Initial attempts in creating explanations for the correct answers to elementary science exam questions by Jansen et al. (2017) addressed answer extraction and explanation creation as a joint task. They like Paul and Frank (2019) extract a short linguistically highlighted path based on their algorithm features through a number of textual knowledge bases such as study guides and science dictionaries, and examine the facts in the traversed path as valid explanation candidates or not. This early approach adopted an open world assumption of generating explanations for a QA pair where the knowledge store for explanation candidate facts was not fixed. Thus, any natural language text or book or even the web could be thought of as a source of candidate facts for explanations. This meant that explanation generation itself could not be realistically quantitatively evaluated across systems in such settings.

In contrast, the newly introduced WorldTree corpus (2018) adopted a closed-world assumption with QA pair explanations defined in terms of human-coded facts, similar to human-coded knowledge for declarative QA such as in LifeCycleQA (Mitra et al. 2019), or in the form of rules encoding the fundamental assumption about puzzles in the puzzle solving domain (Mitra and Baral 2015). In other words, explanation knowledge became confined to a fixed smaller set of about 4,700 human-annotated facts. This facilitated quantitative evaluations for approaches on the task based on the availability of a benchmark human-annotated test set. Thus far, via the TextGraph-19 Explanation Regeneration Shared Task (Jansen and Ustalov 2019), four known systems (Das et al. 2019; Chia et al. 2019; Banerjee 2019; D'Souza et al. 2019) have been developed on the WorldTree corpus. They demonstrated a diverse range of performances from 56.3% *mAP* (Mean Average Precision) to 39.4% *mAP*. Systems by Das et al. (2019), Chia et al. (2019), and Banerjee (2019) had two facets in common: BERT-based neural models; and a reranking paradigm, i.e. the tablestore of facts were first ordered by one approach, following which the ordered facts were then reranked by a different approach. Our system (D'Souza et al. 2019) differed from the rest in that it leveraged a non-neural-network machine learning paradigm. In it, the core system was a traditional hand-crafted feature-based SVM ranker. It was integrated in a hybrid framework, that additionally employed a set of rules to correct for obvious machine learning prediction errors. The features in the machine learning system were mostly generically-oriented, including knowledge sources such as Wiktionary categories and page titles, and FrameNet predicates and arguments (Swayamdipta et al. 2017). The generic features may have been a system limitation since they most likely functioned as undesirable distractors in the machine learning task causing low system performance.

Thus, while our hybrid system at 39.4% was significantly better than the basic TF-IDF system at 29.6%, and on par with Banerjee (2019)'s BERT-based system at 41.3%, it showed much lower performance compared to the top-ranked BERT-based ensemble at 56.3%.

In the new version of our system described in this paper, we re-engineer our earlier system (2019) on two criteria: 1) we train a more effective learning technique as pointwise predictions from an SVM regressor for the task, differing from our earlier pairwise learning-to-rank approach; and 2) we replace the generic set of features with a targeted set of knowledge sources for the task dropping information sources that did not prove informative (e.g., Wiktionary features and predicate-argument frame features); and adding domain-targeted sources such as IR-based optimized TF-IDF facts' rankings, multihop inference targeted features, and BERT-based semantic abstraction features. This system achieves >10 points boost in *mAP* over our shared task system at 53.2% on the development set and 50.7% on the test set. Further, even the learning-to-rank system attains a 3 points boost with better task-specific set of features at 45.9% and 43.3% on the development and test sets, respectively, without having to rely on rules. Our new approach outperforms all existing BERT-based (Devlin et al. 2018) neural systems except the top-ranked computationally intensive approach by Das et al. (2019) which we describe in detail in the experimental section (see Section 7.2).

Finally, we conclude our discussion on the related work by situating our system in the context of two different extended tasks: 1) to implement feedback between explanation generation and question answering to mutually improve the tasks' performances; and 2) to compose knowledge from facts.

***Answering Multi-Choice Questions by Ranking Supporting Facts***.   Pirtoaca et al. (2019) leverage Wikipedia as a source of supporting facts to improve question answering on the ARC corpus. Thus, they considered Wikipedia as the facts knowledge base. Their system introduced a self-attention based neural network that latently learns to rank sentences by their importance related to a given question, whilst optimizing the objective of predicting the correct answer. Their work, which was performed independently of the release of the WorldTree corpus (2018), took the open-world assumption for Wikipedia sources of supporting factual evidence. The release of the human-annotated explanation facts in the WorldTree corpus, however, alleviates such settings in which no supporting facts are explicitly provided and, being specifically human designed, could potentially boost QA performance better. While in our system described in this paper, we are solely addressing the problem of explanation fact ranking and not facts ranking to improve multiple-choice question answering, the latter task is relegated as follow-up work.

***Knowledge Composition from Facts***.   Khot et al. (2020) define the Question Answering via Sentence Composition (QASC) corpus, in which they present for the first time the knowledge composition task. E.g., combining two different facts, i.e. "Differential heating of air produces wind" and "Wind is used for producing electricity" into a single knowledge sentence "Differential heating of air can be harnessed for electricity production ." The QASC corpus involves both elementary and middle-school science domains. A small subset of the composition facts were taken from the WorldTree corpus (2018). This new task is considered useful since such composed knowledge was shown to be a useful signal to boost QA performance (Khot et al. 2020).

Essentially, IR systems on this dataset would need to introduce new concepts or relations in order to discover relevant facts. Further, they must then learn to identify valid compositions of these retrieved facts using commonsense reasoning—functions that are ingrained in our system. Thus, our system, with some task-specific modifications, could also be leveraged for facts composition as future work.

This concludes our discussion on related work. In the next section, we provide details of our approach w.r.t. the task in the WorldTree corpus (2018).

## 5. Approach

As described in Section 3, explanation regeneration for Elementary Science QA pairs is posited as a ranking task given a collection of candidate facts, where for each QA pair explanation, the number of valid facts can vary up to 21 and the desired result is to have all the valid facts top-ranked. Formally, let $(q, ca, f)$ be a triplet consisting of a question $q$, its correct answer $ca$, and a candidate explanation fact $f$ that is a valid or invalid candidate from the given unordered facts tablestore $F_{uno}$. Our task is, for each $(q, ca)$ given $F_{uno}$, to rank the generated $(q, ca, f)$ triplets such that the group $(q, ca, f^c)$ is top-ranked to produce an ordered tablestore $F_o$, where $f^c$ stands for the group of relevant facts in the explanation and $f^c \subseteq F_{uno}$.

Within a preference-based object ordering formalism (Melnikov et al. 2016), the candidate facts $F_{uno}$ comprise the reference set of objects. Training data consists of a set of rankings $\{O_1, ..., O_N\}$ of facts for $N$ $(q, ca)$ training instances, respectively, where for $(q, ca)_i$, the ordering is:

$$O_i : f_a^c \succ f_b^c \succ ... \succ f_g^c \tag{1}$$

such that $O_i$ is an ordering of only the valid facts $f_i^c$ for a $(q, ca)_i$ instance where $|O_i| < |F_{uno}|$. The order relation $\succ$ is interpreted in terms of preferences, i.e., $f_a \succ f_b$ suggests that $f_a$ is preferred to $f_b$ in terms of logical discourse. And the remaining $F_{uno} \setminus f_i^c$ are assigned a uniform least rank.

The next natural question is which functions do we choose to learn the set of orderings for $(q, ca)$ pairs. In particular, two such approaches are prevailing in the literature. The first one reduces the original ordering problem to *regression*: it seeks a model that assigns appropriate scores to individual items and hence is referred to as the *pointwise* approach. The second idea reduces the problem to *binary classification*; the focus is on pairs of items, which is why the approach is also called the *pairwise* approach. Next, we briefly introduce these models in the context of the support vector machine (SVM) class of algorithms and describe how we train them.

At a high-level, the objective of the SVM is to find the optimal separating hyperplane in an N-dimensional space (where "N" is the number of features) which maximizes the margin of classification error on the training data. The margin is defined in terms of certain select training data points that influence the position and the orientation of the hyperplane such that it is at maximal separating distance from the data points in the various classes. These points then constitute the support vectors of the trained SVM. The support vectors lie on boundary lines that run parallel to the classification hyperplane but at the maximal computable distance. Obtaining a maximal margin produces a more generalizable classifier to unseen data instances. Note also that in real-world problems, the boundary lines are more practically considered soft boundaries with an error allowance defined by a slack variable $\xi$, that allows classifications to fall somewhere within the boundary margin from the classification hyperplane. Formally, as an optimization problem, the SVM classification objective is to:

$$\begin{aligned} \min_{w,b,\xi} \quad & \frac{1}{2} w^T w + C \sum_{i=1}^{N} \xi_i \\ \text{s.t.} \quad & y_i - w \cdot \phi(x_i) - b \leq \xi_i \\ & w \cdot \phi(x_i) + b - y_i \leq \xi_i \\ & \xi_i \geq 0 \end{aligned} \tag{2}$$

where $i = 1, ..., N$ for N training instances, $\phi$ is a feature transformation function for input $x_i$, $w$ is the features' weight vector over all instances, and $y_i$ is either +1 or -1. The constant $C > 0$ determines the tradeoff between the norm of the weight vector and error margin defined by slack variable $\xi$.

### *5.1 Pairwise Learning-to-Rank (LTR) for Preference Ordering*

The next question is how can our preference ordering problem be formulated in terms of binary classification. This is possible by the pairwise LTR transformation. Vaguely, this is done by modeling: 1) whether a candidate fact is a valid candidate or not; and 2) for the collection of valid explanation facts, the logical precedence of one fact over another. Thus, these decisions are identified in a relative sense, that is to say, by determining the pairwise preferences between facts in the explanation compared w.r.t. each other and w.r.t. the remaining facts in the tablestore.
Our dataset originally is:

$$S = \{x_{ij}, y_{ij}\} \text{ where } x_{ij} = \phi((q_i, ca_i), f_j)$$

$(q_i, ca_i)$ is the i-th QA pair instance, $f_j$ is the j-th explanation fact from the tablestore where the ordering between facts is known during training and is unknown during development and testing. $\phi$ is a feature transformation function, and $y_{ij} \in \{1, 2, 3, ...K\}$ denotes an order between the $(q_i, ca_i)$ pair and the explanation fact $f_j$ as a graded order w.r.t. the other relevant and irrelevant explanation fact candidates.
By the pairwise LTR transformation, our original dataset $S$ then becomes:

$$S' = \{(x_{ij} - x_{il}), (y_{ij} \theta y_{il})\}$$

where $\theta$ is the rank difference so that $(y_{ij} \theta y_{il}) = 1$ if $y_{ij} > y_{il}$ and -1 otherwise, resulting as a binary classification task. The goal of the LTR algorithm is to acquire a ranker that minimizes the number of violations of pairwise rankings provided in the training set which is attempted as the above classification problem.

Essentially, since pairwise LTR only considers the labels where $y_{ij} > y_{il}$ between relevant candidates and $y_{il} > y_{ij}$ between relevant and irrelevant candidate pairs, respectively, we assign (a) higher ranks to the relevant facts to indicate precedence, and (b) equal ranks of 1 to all irrelevant facts. Thus, for consecutive relevant instances we offset the ranks by 1. That is, for a given $x_i$, if there are $n_i$ correct facts in the explanation, then the first fact in the correct ordering receives a rank of $|n_i| + 1$, the second is ranked as $|n_i|$, and so on; the last correct fact receives rank 2, and all irrelevant facts have rank 1.

### *5.1.1 Training LTR for QA Pair Explanation Fact(s) Preference Ordering*

We use the SVM LTR learning algorithm as implemented in the SVM$^{rank}$ software package (Joachims 2006). To optimize ranker performance, we tune the regularization parameter C (which establishes the balance between generalizing and overfitting the ranker model to the training data). However, we noticed that a ranker trained on the entire tablestore set of facts is not able to learn a meaningful discriminative model at all owing to the large bias in the negative examples outweighing the positive examples (consider that the number of relevant explanation facts range between 1 to 21 whereas there are 4,789 available candidate facts in the tablestore). To overcome the class imbalance, we tune an additional parameter: the number of negative facts for training. Every $(q, ca)$ training instance is assigned 1000 randomly selected irrelevant explanation facts. We then tune the selection of the number of irrelevant explanation facts ranging between 500 to 1,000 in increments of 100.

Both the regularization parameter and the number of negative explanation facts are tuned to maximize performance on development data. Note, however, that our development data is created to emulate the testing scenario. So every $(q, ca)$ instance during development is given all 4,789 facts to obtain results for the overall ordering task.

*5.1.2 Testing LTR for QA Pair Explanation Fact(s) Preference Ordering*

During testing time, all facts in the tablestore are respectively paired with a $(q, ca)$ test instance. Each instance is then represented by the features defined in our system before being fed as input to the SVM-LTR trained model. The trained model then predicts ranking scores for each data instance, where the scores are then used to order the facts as the regenerated explanation for the given $(q, ca)$ test instance.

### *5.2 Pointwise Preference Ordering by Regression*

SVM regression differs from the SVM classification objective in that instead of optimizing over binary targets, the optimization is performed for real-valued targets. To facilitate this, regression is then defined in terms of an $\varepsilon$-precision objective. In other words, we do not care about training errors as long as they are less than $\varepsilon$. Further, as in the classification objective with soft decision boundaries, similar allowances are made with slack variables in the regression context, but defined over targeted regression precision. Formally, the regression optimization problem is defined as follows:

$$\begin{aligned}
\min_{w,b,\xi,\xi^*} \quad & \frac{1}{2}w^T w + C \sum_{i=1}^{N} (\xi_i + \xi_i^*) \\
\text{s.t.} \quad & y_i - w \cdot \phi(x_i) - b \leq \varepsilon + \xi_i \\
& w \cdot \phi(x_i) + b - y_i \leq \varepsilon + \xi_i^* \\
& \xi_i, \xi_i^* \geq 0
\end{aligned} \quad (3)$$

where, $i = 1, ..., N$ for N training instances, $\phi$ is a feature transformation function for input $x_i$, $w$ is the features' weight vector over all instances, and $y_i$ is a real-valued target, and $\varepsilon$ is the regression targeted precision. The constant $C > 0$ determines the tradeoff between the norm of the weight vector and error margin defined by slack variables $\xi, \xi^*$.

Next, an important question is, how to represent our ordering problem in terms of a regression objective. We do this by defining regression targets in terms of the preference ordering expectations (Kamishima et al. 2005 2010) rather than true regression quantifications. In our dataset $S = \{x_{ij}, y_{ij}\}$ where $x_{ij} = \phi((q_i, ca_i), f_j)$, the label $y_{ij}$ for the correct candidate explanation facts are indicated as unit graded relevance in order of their preference, while all the incorrect candidates are relegated to a uniform least rank. The facts are assigned ranks in exactly the same manner as the pairwise LTR setting.

By using regression for the preference ordering of facts in explanations, we make the assumption that all facts can be treated independently w.r.t. each other. Such assumptions are highly contingent on the properties of the underlying dataset and may not apply in all preference ordering or ranking scenarios. In contrast to the regression setting, the pairwise LTR is, in principle, applicable in any ordering scenario. Evidently, in the WorldTree corpus (2018), the order of facts in the explanations are based on logical precedence rather than discourse, wherein the discourse linguistic cues are not readily apparent in most explanations (consider the example depicted in Table 1). Thus, from the perspective that considers a pure logical precedence between the explanations' facts, a regression setting that predicts logical precedence weights as target values for the facts is nonetheless relevant and a sound modeling of the task.

*5.2.1 Training SVR for QA Pair Explanation Fact(s) Preference Ordering*

We use the SVR learning algorithm (Vapnik 1995) as implemented in the SVM$^{light}$ software package (Joachims 2002) (hence called SVM$^{reg}$ since we employ its regression setting). Similar to the ranker system, to optimize regression performance, we tune the regularization parameter C on the development set with all the other parameters at their default values. Again like in the ranking training setup, we randomly select a smaller set of irrelevant explanation facts to learn a meaningful discriminative model which are tuned on the development set to range between 500 to 1,000 in increments of 100.

Note that our development data is created as usual to emulate the testing scenario given $F_{uno}$. So every QA pair instance during development is given all 4,789 candidate facts for regression predictions.

*5.2.2 Testing SVR for QA Pair Explanation Fact(s) Preference Ordering*
The testing scenario is identical to the pairwise LTR model w.r.t. data input instance generation. In this case, however, a trained SVR model predicts regression target values which are then used to rank the explanation facts.

## 6. Features for the Explanation Regeneration Task

In this section, we elaborate on our feature function $\phi$ introduced with our formal models used to transform a $(q, ca, f)$ triplet to a one-hot encoded feature vector $x$ as data instances for the learning algorithms.

Our motivation in selecting these features was to encode linguistically the necessary world and commonsense knowledge required for unifying facts as explanations to Elementary Science QA. There are six main feature groups which are described next.

### 6.1 Bags of lexical features (70,949 total features)

This feature group most generically encodes the lexical overlap criteria by including features as lemmas of $q/ca/f$; lemmas shared by $q$ and $f$, $ca$ and $f$, and $q$, $ca$ and $f$; 5-, 4-, and 3-gram prefixes and suffixes of $q/ca/f$; 5-, 4-, and 3-gram prefixes and suffixes shared by $q$, $ca$, and $f$; and $f$'s table type from the provided annotated tablestore data. The lemma and n-gram features are filtered for common pronoun and prepositional stop words.

### 6.2 ConceptNet (294,249 total features)

We hypothesize that semantic features, in particular commonsense knowledge, could be useful for the explanations to elementary science QA pairs. Since elementary science questions query general knowledge about common nouns like animals, planets, occupations, etc., we find that ConceptNet (Speer et al. 2017) as a resource with its focus on the general meanings of all words, whether they be nouns, verbs, adjectives, or adverbs, and less on named entities, is perfectly suited to our task. Let us illustrate with an example.

**Question** Which animal eats only plants?
**Answer** Rabbit
**Explanation**
herbivores only eat plants;
a rabbit is a kind of herbivore;
a rabbit is a kind of animal;
**ConceptNet conceptualizations for "rabbit"**
animal, herbivore

In this example, ConceptNet tells us that the answer "rabbit" is an "animal" and a "herbivore", among other things. Extending the answer with this knowledge enables better semantic connection between the $q$, $ca$, and all three explanation facts, in the absence of which, the ranking could experience a semantic drift toward irrelevant explanation facts such as "long ears are a part of a rabbit" or "a jackrabbit is a kind of rabbit".

Given the potential usefulness of ConceptNet for our task, we create conceptualization features as follows. The top 50 conceptualizations of $q/ca/f$ words; top 50 conceptualizations shared by $q$ and $f$, $ca$ and $f$, and $q$, $ca$ and $f$ words; and the relation names that $q/ca/f$ words as ConceptNet facts participate in such as FormOf, IsA, HasContext, etc. E.g., for word 'tea' in $q/ca/f$, the ConceptNet facts are 'tea ReceivesAction brewed', 'tea HasA caffeine', 'teaIsA beverage', etc., from which the features are 'ReceivesAction_brewed', 'HasA_caffeine', and 'IsA_beverage'.

Note, since ConceptNet returns an ordered list of conceptualizations for queried terms ordered w.r.t. precision as most precise to generic, therefore, by selecting only the top 50 for the first two features, we control for genericity of the conceptualizations in the features considered.

### *6.3 OpenIE relations (36,989 total features)*

We introduce features computed as open information extraction relation triples using the OpenIE tool (Angeli et al. 2015). We observed that within the triple representations of the question, correct answer, and explanation fact sentences were the content words that were needed to link across other feature groups (e.g., ConceptNet). This linkage also enabled indirect connections between $(q, ca)$ and $f$. Let us illustrate with an example:

**Question** Which of the following properties provides the BEST way to identify a mineral?

**Answer** Hardness

**Explanation**

hardness is a property of a material or an object and includes ordered values of malleable or rigid;

In the example, the given fact is top-ranked in the explanation. For it, from OpenIE we get the relation triple (hardness → is a property of → material). Further, ConceptNet tells us that the answer Hardness is related to concepts "property," "material property," etc. We see how pooling these information units together enables a unified word cloud involving the question, correct answer, and explanation fact for the terms "hardness," "property," and "material." Features that enable grounding externally computed terms to the lexical items given in the QA pair or explanation facts create a tighter overlap improving task performance.

Given the potential usefulness of inter-sentence OpenIE triples for explanation generation, we create features as follows. For each triple produced by the parser, the features are: the $q/ca/f$ lemmas in the relation subject role, shared $q$, $ca$, and $f$ subject lemmas, $q/ca/f$ lemmas in the relation object role, shared $q$, $ca$, and $f$ object lemmas, and $q/ca/f$ lemma as the relation predicate.

### *6.4 Multihop inference specific features (2,620 total features)*

These features are a more selective bag of lexical features for obtaining matches with a positional emphasis. We find that adding positional information for lexical matches is a useful heuristic to identify the concepts that are the focus of the $(q, ca)$ and explanation facts. Consider the underlined words in the two subsequent examples in this section.

As shown in the examples, often the focus word of the $(q, ca)$ are at the start or end and also at the start and end of the $f$. Further, for one- or two-word $ca$, we can directly infer it as a focus concept, in which case we try to find a match with $f$ where they are the first or last word. And for focus words that are verbs, they tend to occur in the middle.

**Question** There are different types of desert. What do they all have in common?
**Answer** low rainfall
**Explanation**
a desert environment has low rainfall

**Question** Sonar helps people find which information about an object?
**Answer** Location
**Explanation**
sonar is used to find the location of an object;
the location of an object can be used to describe that object;

The following features are considered in this group: length of *q* and *ca*; positions of *q/ca* verbs in the phrase (as 0 if it is the first word, 1 if it is the second word, and so on); of the verbs shared by *q* and *f* do they occur among the first few words or middle or last words.[b]; if *ca* is a uni- or bigram, does *f* contain all its words/lemmas?; does *f* contain the last *q* lemma/word?; is the last *q* lemma/word in the first position of *f*?, is it in the last position of *f*?; is the first *q* lemma/word in the first position of *f*?

Notably, the positional emphasis observations made for the features in this category are contingent on the language of the corpus. E.g., we observe verbs as focus words occurring in the middle of the sentence because the WorldTree corpus is annotated for English sentences which follow the subject-verb-object (SVO) sentence structure. Thus, it is necessary to highlight that if the WorldTree corpus was annotated for a language with a different sentence structure, e.g., Persian which follows the SOV order, the verbs would be expected at the end of the sentence and our 'multihop inference specific features' would need to be accordingly adapted. Further, since the WorldTree corpus handles the Elementary Science level, it contains fairly short atomic sentences and this accounts for us finding focus nouns at specific locations as the sentence start or end and the focus verbs in the middle. A worthwhile consideration then is that for higher scientific levels the positions of the focus words could be different and harder to pin down as a fixed location as was possible in our case for the WorldTree corpus.

### *6.5 TF-IDF Ranking (750,283 total features)*

The explanation regeneration task performance via ranking based on cosine similarities between TF-IDF weighted $(q, ca)$ appended text and each fact candidate proves surprisingly effective for the task (see scores in Evaluation section). We use the *TF-IDF Iterated* variant by Chia et al. (2019) to encode the text. The ranks obtained by cosine similarity on these instances are then used as features for the SVM learner. We hypothesize that employing the TF-IDF-based cosine similarity ranks as features will provide a baseline ordering signal to the learning algorithm.

Our TF-IDF features per $(q, ca, f)$ are the following: *f*'s rank; *f*'s binned rank in bins of 50; *f*'s binned rank in bins of 100; whether *f* is in top 100 or 500 or 1000?

### *6.6 BERT embeddings*

BERT-based (Devlin et al. 2018) context embeddings are our last features category. The out-of-box BERT model is pretrained on millions of words from Wikipedia which as a commonsense knowledge source is already pertinent to elementary science QA. Thus, we simply query the BERT embeddings from the pretrained model using the bert-as-a-service library. Thus, for each data instance word, we extract their BERT embedding features that can easily can be combined with the other linguistic features. This can be viewed as a semantic projection of an elementary science concept in the Wikipedia encyclopedia space. Specifically, we query the BERT$_{Base}$ UNCASED ENGLISH model: 12 layers, 768 hidden units, 12-heads, with 110M parameters that outputs a 768 dimensional vector for a given input text. We treat each dimension of this context vector as a separate feature for representing the instance.

---

[b]For first, middle, and last words, using a window 1/4 the size of the total words, centered on the middle, we find the middle portion of the sentence, at its LHS, the first portion, and at the RHS, the last portion of the sentence.

While the earlier five feature categories enabled extending the $(q, ca, f)$ vocabulary beyond the given words both lexically and conceptually, with BERT embeddings we aim to leverage semantic abstractions as features. We hypothesize such features would be useful in creating semantic associations between the elements in the $(q, ca, f)$ triple which are topically similar based on knowledge from Wikipedia. As in the following example.

**Question** Diamonds are formed when carbon is placed under extreme heat and pressure. This process occurs
**Answer** beneath the surface of Earth.
**Explanation**
the formation of rock is a kind of process;
diamond is a kind of mineral;
rock is made of minerals;
the formation of diamonds occurs beneath the surface of the Earth by carbon being heated and pressured

In the example, considering the focus words "diamonds," "earth," and "minerals" that reflect the topics of the QA pair, the word "minerals" in the fact is neither present in the $q$ or $ca$, but is relevant to the semantic topic of the $(q, ca)$. We hypothesize that BERT features will help capture such topicalized semantic abstractions of similarity.

We tested two ways of obtaining BERT features for $(q, ca, f)$ triples: i) query BERT separately for the question, correct answer, and fact embeddings, respectively, obtaining three 768 dimensional feature sets and resulting in 2,304 additional features from BERT per instance; and ii) query BERT for aggregate 768-dimensional embedding features for the $(q, ca, f)$ triple. In this configuration, we use the special [SEP] token to demarcate the $q$, $ca$ and $f$ segments. The start-of-the-sequence [CLS] token then learns the encoding for the entire input sequence. Experiments indicated that the latter method is a better-suited representation for the task, while the former method is ineffective. This can be attributed to two reasons: (1) The second approach allows the model to capture not only the representations of the sequences but also the proximity between the input segments. Thus, the [CLS] token in this case is a latent similarity vector for the $q$, $ca$, and $f$. In the first case, it merely represented the sequences separately. And, (2) the length of the vectors produced in case (i) is three times the length of the vectors produced in case (ii). This results in sparse features which in turn limits the performance of the SVM classifier.

Thus all six feature categories used to represent $(q, ca, f)$ triples, when taken together, should readily address the multi-step inference process between $(q, ca)$ and $f$ candidates. This is since we have features extending the given information in the $(q, ca)$ with world knowledge generically (e.g., ConceptNet, BERT) with other features providing lexical glue (at generic and task-specific levels) enabling traversing the $(q, ca, f)$ via multiple hops (e.g., multihop inference lexical features, OpenIE relations). Therefore, for themes such as about vehicles, as an example from Jansen et al. (2016), describing its mechanisms, its purpose, its needs, and its functions, our various feature groups can take into account such diverse aspects of the real world.

Finally, in Figure 3 we depict our overall approach including the feature modules and the two SVM-based machine learners we employed.

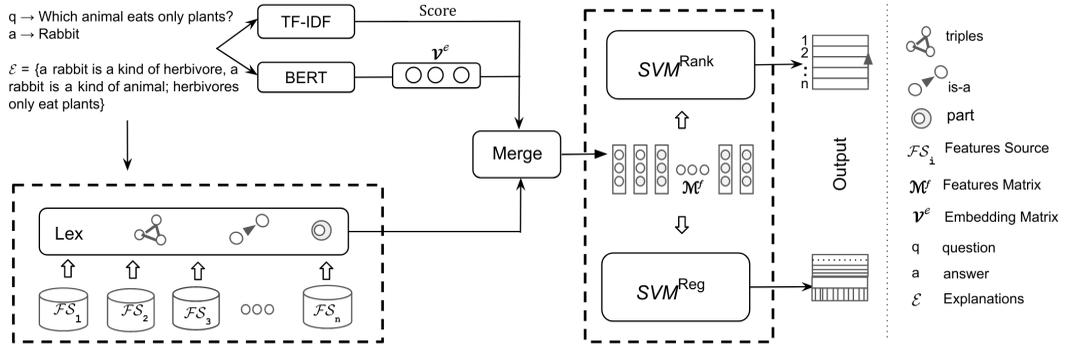

Figure 3: Overall representation of our approach. For representing $(q, ca, f)$ triples, the feature categories used include Lex (lexical features), TF-IDF (information retrieval features), and BERT-based features among others. The data instances are then represented as a features matrix, separately as training data, development data, and testing data. Two variations of the SVM algorithm ($SVM^{Rank}$ and $SVM^{Reg}$) are then used to learn Explanation Regeneration models.

## 7. Evaluation

### 7.1 Experimental Setup

*Dataset.* The experimental corpus of this study is the Worldtree corpus (2018), introduced in detail in Section 2. In our experiments, we maintain the same dataset fold splits as provided by the dataset creator.

*Evaluation Metrics.* We report one set of results in terms of the mean average precision *mAP* metric which is a standard in IR ranking tasks. With the *mAP* metric score we see to what extent our system returns the relevant explanation facts as top-ranked. To evaluate our system for ordering the relevant explanation facts w.r.t. each other, we employ the Precision@k and Recall@k metrics, where k is the group of top-ranked facts ranging between 2 to 50 facts in increments of 2. Note that for these latter metrics, a correct score is considered for each fact returned at its exact given position in the ordered facts in the explanation. For instance, to get a score of 1.0 by the Precision@2 metric, the predicted top-scoring fact should be the top-ranked fact and the predicted second top-scoring fact should be the second ranked fact in the gold data. These latter evaluations are a closer test of our system as satisfying the task defined in the WorldTree corpus (2018), in other words, to see if it logically orders the relevant facts at all.

*Parameter Tuning.* For the $SVM^{rank}$ and the $SVM^{reg}$ systems, we jointly tune the C and the number of negative training instances parameters on development data. Our best $SVM^{rank}$ model when evaluated on development data was obtained with C = 0.8 and 1000 negative training instances, while our best $SVM^{reg}$ model was obtained with C = 0.005 and 900 negative training instances.[c]

We compare our models with nine existing systems as reference performances, where the systems we compare with have varying degrees of complexity from simple IR approaches to neural-based machine learning approaches. Further, as we show in the results eventually, the systems also have varying degrees of performances not necessarily correlated with the system complexity—TF-IDF approaches prove surprisingly effective on this task. In the following section, the nine systems we evaluate against are briefly described.

---

[c] For parameter tuning, C is chosen from the set {0.005,0.05,0.1, 1,10,50,100} and the number of negative training instances is chosen from the set {500,600,700,800,900,1000}.

### 7.2 Nine Reference Evaluations

***TF-IDF Baseline.*** Facts are ranked by cosine similarity of their TF-IDF representation with the TF-IDF representation of the query string composed of the question and all the available answer choices.

***TF-IDF Baseline features + SVM$^{rank}$ (Jansen and Ustalov 2019).*** For each data instance, two cosine similarity scores were computed: one between the question TF-IDF vector and the candidate explanation fact vector; and another between the correct answer and the candidate fact vectors. These scores were used as features within an SVM$^{rank}$ (Joachims 2006) setting and a ranked list of facts were predicted.

***Generic Feature-rich SVM$^{rank}$ (D'Souza et al. 2019).*** Our previous system had five main feature categories including OpenIE (Angeli et al. 2015), ConceptNet (Liu and Singh 2004), Wiktionary, and FrameNet (Swayamdipta et al. 2017) representations for each $(q, ca, f)$ triple are employed, which are then ranked by SVM$^{rank}$ (Joachims 2006).

***Rules + Generic Feature-rich SVM$^{rank}$ (D'Souza et al. 2019).*** In this hybrid model, the *Generic Feature-rich SVM$^{rank}$* system output is corrected for obvious errors by a set of 11 re-ranking rules applied sequentially, pipelined to the SVM system output. As an example of a rule consider: all facts that contain the bigram or unigram correct answer word are to be top-ranked.

***BERT Iterative Re-ranking (Banerjee 2019).*** The system models explanation regeneration using a re-ranking paradigm, where BERT (Devlin et al. 2018) transformer models are used to provide an initial ranking, and the top-15 facts output by the BERT model are re-ranked using a custom-designed relevance ranker to improve overall performance. Note that in comparison with the top-performing BERT-based model (Das et al. 2019), described last in this section, this system is run in a BERT out-of-box configuration.

***Optimized TF-IDF (Chia et al. 2019).*** This system differs from *TF-IDF Baseline* in the following ways: all incorrect answer choices are dropped from the query string; the query and the fact strings are additionally preprocessed by lemmatization and the removal of their stopwords.

***Iterated TF-IDF (Chia et al. 2019).*** Where in the *Optimized TF-IDF* system, the query string consists of only the question and the correct answer, in this system, the query string is iteratively expanded to include the top-ranked fact. After each expansion step, cosine similarity is rerun on the remaining facts to obtain the next top-ranked fact. This process is iteratively repeated until all facts are ranked.

***BERT Re-ranking with Iterated TF-IDF scores (Chia et al. 2019).*** A BERT regression module is trained to predict the relevance score for each $(q + ca, f)$ pair, where the relevance score is the *Iterated TF-IDF* system rankings; and in the interest of lowering the computational complexity and runtime, the model is trained and tested to rerank only the top 64 of the *Iterated TF-IDF* system output.

***BERT Re-ranking with inference chains (Das et al. 2019).*** It is an ensemble model composed of BERT-based path ranker and a more advanced reranking system (Nogueira and Cho 2019) which they employ as a ranker. The BERT-based path ranker uses a sophisticated multi-step design. The initial step involves obtaining the top 50 facts based on TF-IDF similarity with $(q, ca)$ query. In the next step, 1-hop lexical similarity paths are traced from each fact in the retrieved 50 facts list to the remaining facts in the tablestore. Finally, the BERT path ranker is trained on pairwise fact instances. Instances are formed by exhaustively pairing each fact in the top 50 TF-IDF list with all the corresponding retreived facts at a 1-hop lexical distance from it such that the pairs where both facts constitute the explanation for the given query are a true instance for the BERT path ranker, and others are false. The overall ensemble system then relies on this BERT-based path ranker output for a score threshold of above 0.5, else it uses a reranker (Nogueira and Cho 2019).

With its chaining of facts, this system models a vital aspect of the corpus: i.e., some valid explanation facts directly lexically overlap with the QA pair, and others lexically overlap with other valid facts. As we will see next, this system has the overall best performance, and is the only system we do not outperform. We note, however, that it also presents a high degree of computational complexity. Consider in our system, during training, each QA pair is linked with only roughly 1000 explanation facts; whereas the Das et al. system, would construct training instances as follows: for each QA pair, given top-50 ranked facts, assuming each fact has a 1-hop chain to at most 200 other facts, this would result in nearly 10,000 instances. A larger training dataset generally implies a larger training time - a fact that is particularly true in the case of BERT models as the Das et al. system, while also true for SVMs, although in the SVM case, the number of features would also matter. Further, in the test scenario, while the Das et al. system would still evaluate for 10,000+ odd chains, we would merely check for all the facts in tablestore which presently is about 5,000. Thus, the Das et al. (2019) system is the most effective and at the same time the most computationally intensive of all the systems including ours.

### 7.3 Results and Discussion

Table 6 shows the elementary science QA pair explanation fact preference ordering results in terms of *mAP* with best results from the reference system and ours (last two rows) in bold.

Between our models, we find that $SVM^{reg}$ is significantly better[d] than $SVM^{rank}$ by applying the paired *t*-test to their adjacent scores in the table. Thus, given our underlying dataset, a pointwise learning approach proves better suited to it than a pairwise learning approach. Nevertheless, the latter is still a valid model, in principle, for the task as it does not rely on the strong, and seeming unrealistic, independence assumption between instances made by the $SVM^{reg}$ model. However, since $SVM^{reg}$ significantly outperforms $SVM^{rank}$ at $p < 0.05$, it proves practically better suited on this dataset implying that the non-independence assumption between facts by $SVM^{rank}$ is not a crucial factor in learning the task defined in the data.

Compared with the nine reference systems, our $SVM^{reg}$ approach significantly outperforms eight of the models. This set of systems also includes the neural ranking model by Banerjee (2019), which our system surpasses by +9.4/+10.9 points in *mAP*; as well as a neural re-ranking model by Chia et al. (2019) which we surpass by +3 *mAP*. Although we observe lower performance when compared to the best-performing approach (-5.6/-5.3) by Das et al. (2019), theirs is a significantly computationally complex system than ours as explained earlier (Section 7.2). Finally, in terms of scalability, next to our feature-rich $SVM^{reg}$ are the *Iterated TF-IDF* or *Optimized TF-IDF* models by Chia et al. (2019). Where the simplistic *Optimized TF-IDF* system, as expected, significantly underperforms feature-rich systems, nevertheless its ranking output as features in our system proves effective as we will see in the ablation analysis results (Section 7.3.1).

With our re-engineered system leveraging domain-targeted features, we have significantly outperformed our earlier system that was based on generic linguistic features (2019). Our $SVM^{rank}$ is at +9.2/+8.8 compared to the system without rules and at +3.9/+1.5 compared to the hybrid system; and our $SVM^{reg}$ is at +16.6/+16.1 to the without-rules system and +11.3/+8.8 to the hybrid system. Thus, we see in contrast the task impact obtained from an effective learning algorithm and a set of features that specifically models the domain in our new system version.

In the preceding paragraphs, we have discussed the performance of our system for producing top-ranked relevant facts. Next, we briefly examine the performance of our system for ordering relevant facts w.r.t. each other among the top-ranked. This is presented by the Precision@k and Recall@k evaluations depicted in Figure 4. We see that at low values of k (i.e., $\leq 26$), the ordering performance of $SVM^{reg}$ is distinctly better than $SVM^{rank}$ in terms of both precision and recall. However, beyond 26 facts there is no evident difference between them. At k = 2, $SVM^{reg}$ has a recall rate of ~0.42 at a precision rate of roughly ~0.38. This indicates how often the top two automatically ranked facts are correct and in the top two order. At k = 26, we see that 75% of the facts are retrieved (recall rate of ~0.75), however, at low precision of 16% (precision rate of ~0.16). Part of the low precision score can be attributed to retrieving also the irrelevant facts for explanations with less than 26 facts, nonetheless, these standard metrics are not modifiable for cases when all the relevant facts are already retrieved within the 26 for such explanations. Aside from the spike in the precision score at k = 4 for SVM$^{rank}$, both SVM$^{reg}$ and SVM$^{rank}$ show fairly stable precision and recall rates with steady expected decline and climb rates, respectively.

---

[d]Unless otherwise stated, all statistical significance tests are paired *t*-tests with $p < 0.05$.

| | Approach | mAP Test | mAP Dev |
|---|---|---|---|
| 1 | BERT Re-ranking with inference chains (Das et al. 2019) | **56.3** | **58.5** |
| 2 | BERT Re-ranking with Iterated TF-IDF scores (Chia et al. 2019) | 47.7 | 50.9 |
| 3 | Iterated TF-IDF (Chia et al. 2019) | 45.8 | 49.7 |
| 4 | Optimized TF-IDF (Chia et al. 2019) | 42.7 | 45.8 |
| 5 | BERT iterative re-ranking (Banerjee 2019) | 41.3 | 42.3 |
| 6 | Rules + Generic Feature-rich SVM$^{rank}$ (D'Souza et al. 2019) | 39.4 | 44.4 |
| 7 | Generic Feature-rich SVM$^{rank}$ (D'Souza et al. 2019) | 34.1 | 37.1 |
| 8 | TF-IDF Baseline features + SVM$^{rank}$ (Jansen and Ustalov 2019) | 29.6 | – |
| 9 | TF-IDF Baseline | 24.8 | 24.4 |
| | **Targeted Feature-rich SVM$^{rank}$ (Ours)** | 43.3 | 45.9 |
| | **Targeted Feature-rich SVM$^{reg}$ (Ours)** | **50.7** | **53.2** |

Table 6. : Mean Average Precision (*mAP*) percentage scores for Elementary Science Q&CA Explanation Regeneration by our systems (last two rows) compared with nine reference systems on testing (Test) and development (Dev) datasets, respectively.

As a summary statement for our results presented in this section, we see that at 50.7% *mAP* of ranking the valid facts as top-ranked, only a small proportion of the predictions are in exact order based on the gold standard.

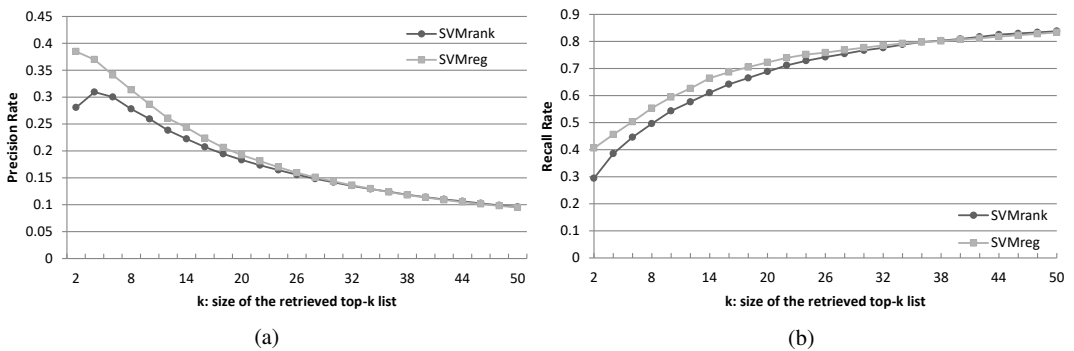

Figure 4: Fact ordering evaluations of our system for top-ranked explanation facts in terms of Precision Rate @ k (subfig. a) and Recall Rate @ k (subfig. b) on the test data.

*7.3.1 Feature Ablation Results*
To provide further insights on the impact of adding different feature groups, we show ablation analysis results in Table 7. Our ablation analysis strategy is to append each of the six feature groups, one at a time, to the baseline features as individual ablation experiments.

| | Feature Type | $\text{SVM}^{rank}$ mAP | | $\text{SVM}^{reg}$ mAP | |
|---|---|---|---|---|---|
| | | Dev | Test | Dev | Test |
| 1 | Bag_of_Lex | 34.01 | 30.44 | 39.57 | 37.43 |
| 2 | Bag_of_Lex+ConceptNet | 36.10 | 32.72 | 41.47 | 39.10 |
| 3 | Bag_of_Lex+OpenIE | 32.61 | 30.59 | 39.83 | 37.12 |
| 4 | Bag_of_Lex+Multihop | 35.33 | 32.46 | 41.90 | 39.43 |
| 5 | Bag_of_Lex+TF-IDF | 40.53 | 38.18 | 51.24 | 46.99 |
| 6 | Bag_of_Lex+BERT | 40.90 | 39.89 | 47.89 | 46.93 |

Table 7. : Ablation results of $\text{SVM}^{rank}$ and $\text{SVM}^{reg}$ on the dev and test sets, respectively, in terms of percentage *mAP* with feature groups (from six feature types considered) added one at a time to the bag of lexical features. TF-IDF and BERT features have highest impact.

From the reported scores in the table, we observe that on both the $\text{SVM}^{rank}$ and $\text{SVM}^{reg}$ learners, respectively, that TF-IDF and BERT features show highest impact; multihop and ConceptNet features were the second most impactful feature types. And the least impact was from the OpenIE features which showed no improvement with $\text{SVM}^{reg}$ and a minor improvement with $\text{SVM}^{rank}$. Nonetheless, we retain this feature group since its ablation analysis doesn't show a negative impact on system performance.

Next, we select a development set example for two of the most impactful feature groups, i.e. ConceptNet and BERT features, for qualitative analysis. The rest of our features can be similarly justified. In both examples, $r_p$ is the predicted rank and $r_g$ is the gold rank.

First, considering the ConceptNet features, a qualitative examination of our results showed that its added commonsense world knowledge prevented semantic drift in several cases. We explain this with the help of the selected example below. In the example, the concepts "tree," "photosynthesis," and "leaves" are the content concepts in the question and the correct answer. These three concepts are also the content concepts in the three gold explanation facts taken together, however, "plant" is an additional content concept in the explanation facts. The task then is to link the facts with the "plant" concept better with the question and the correct answer so that they can be ranked higher. In this respect, from ConceptNet we have the information that the "plant" entity has a class "photosynthetic organism." Our hypothesis was that this additional information should help boost the ranks for the 2nd and the 3rd gold facts. As we can see in the example, adding the ConceptNet feature has indeed helped boost the ranks for the 2nd and 3rd gold explanation facts since it linked "plant" with a focus concept from the question, i.e. "photosynthesis." Further, comparing the **Before** and **After** sets, we can see that the semantic coherence from ConceptNet lower-ranked facts with concepts as "fruit," "eating," "digestion," "animals," "consumers."

In the second example, we qualitatively depict the impact of adding BERT features. We glean the theme of the QA pair as "falling under gravity." While the dotted phrases "gravitational force," "fall," and "falling" encompass the theme, they are not directly present in the $(q, ca)$ pair unlike the underlined phrases. The third fact, i.e. "come down is similar to falling," that contains one of the $(q, ca)$ absent thematic phrases, viz. "falling," after adding BERT features, attains a significant performance boost by 7 ranks shown in **After** ranked collection. We posit this is by the better abstract theme modeled by BERT features. As a consequence, several of the unrelated facts were then ranked lower. Consider the facts such as "rubber is a kind of material," "to bounce back means to reflect," "objects are made of materials or substances or matter" that were no longer intervening in the returned **After** result collection.

**Question** In which part of a <u>tree</u> does <u>photosynthesis</u> most likely take place?
**Answer** <u>leaves</u>
**Before**

*[$r_p$ 1 & $r_g$ 1]* a <u>leaf</u> performs <u>photosynthesis</u> or gas exchange
*[$r_p$ 2 & $r_g$ na]* a leaf is a part of a tree
*[$r_p$ 3 & $r_g$ na]* a leaf absorbs sunlight to perform photosynthesis
*[$r_p$ 4 & $r_g$ na]* fruit is a part of a plant or tree
*[$r_p$ 5 & $r_g$ 2]* a <u>leaf</u> is a part of a green <u>plant</u>
*[$r_p$ 6 & $r_g$ na]* if a leaf falls off of a tree then that leaf is dead
*[$r_p$ 7 & $r_g$ na]* eating or digestion is when an organism takes in nutrients from food into itself by eating
*[$r_p$ 8 & $r_g$ na]* a tree is a kind of living thing
*[$r_p$ 9 & $r_g$ na]* green plants provide food for themselves or animals or consumers by performing photosynthesis
*[$r_p$ 10 & $r_g$ 3]* a <u>tree</u> is a kind of <u>plant</u>

**After**

*[$r_p$ 1 & $r_g$ 1]* a <u>leaf</u> performs <u>photosynthesis</u> or gas exchange
*[$r_p$ 2 & $r_g$ na]* a leaf is a part of a tree
*[$r_p$ 3 & $r_g$ 2]* a <u>leaf</u> is a part of a green <u>plant</u>
*[$r_p$ 4 & $r_g$ na]* take place means happen
*[$r_p$ 5 & $r_g$ na]* fruit is a part of a plant or tree
*[$r_p$ 6 & $r_g$ na]* a leaf absorbs sunlight to perform photosynthesis
*[$r_p$ 7 & $r_g$ 3]* a <u>tree</u> is a kind of <u>plant</u>

**Question** If you bounce a rubber ball on the floor, it goes up and then <u>comes down</u>. What causes the ball to <u>come down</u>?
**Answer** <u>gravity</u>
**Before**

*[$r_p$ 2 & $r_g$ 1]* <u>gravity</u> or <u>gravitational force</u> causes objects that have mass or substances to be pulled down or to <u>fall</u> on a planet
*[$r_p$ 1 & $r_g$ 2]* a ball is a kind of object
*[$r_p$ 3 & $r_g$ na]* the floor is a kind of object
*[$r_p$ 4 & $r_g$ na]* rubber is a kind of material
*[$r_p$ 5 & $r_g$ na]* gravity means gravitational pull or gravitational energy
*[$r_p$ 6 & $r_g$ na]* gravity is a kind of force
*[$r_p$ 7 & $r_g$ na]* to bounce back means to reflect
*[$r_p$ 8 & $r_g$ na]* objects are made of materials or substances or matter
*[$r_p$ 9 & $r_g$ na]* where something comes from is a source of that something
*[$r_p$ 10 & $r_g$ na]* gravity pulls objects towards planets
*[$r_p$ 11 & $r_g$ na]* a container contains objects or material or substances
*[$r_p$ 12 & $r_g$ 3]* <u>come down</u> is similar to <u>falling</u>

**After**

*[$r_p$ 2 & $r_g$ 1]* <u>gravity</u> or <u>gravitational force</u> causes objects that have mass or substances to be pulled down or to <u>fall</u> on a planet
*[$r_p$ 1 & $r_g$ 2]* a ball is a kind of object
*[$r_p$ 3 & $r_g$ na]* the floor is a kind of object
*[$r_p$ 4 & $r_g$ na]* gravity means gravitational pull or gravitational energy
*[$r_p$ 5 & $r_g$ 3]* <u>come down</u> is similar to <u>falling</u>

### 7.3.2 Evaluating SVM$^{reg}$ for Multihop Inference

In this section, we present our best system performance, i.e. the SVM$^{reg}$ system with all six features' categories, for multihop inference as a function of explanation lengths, i.e. the number of

24  Ranking Facts for Explaining Answers to Elementary Science Questionsfacts. Presumably, the more the facts in the explanation the greater the implication of the presence of the multihop phenomenon, i.e. lexical hops across explanation facts to attain shared lexical matches with other explanation facts and the other explanation facts then with the $(q, ca)$ pair. This result is depicted in Figure 5.

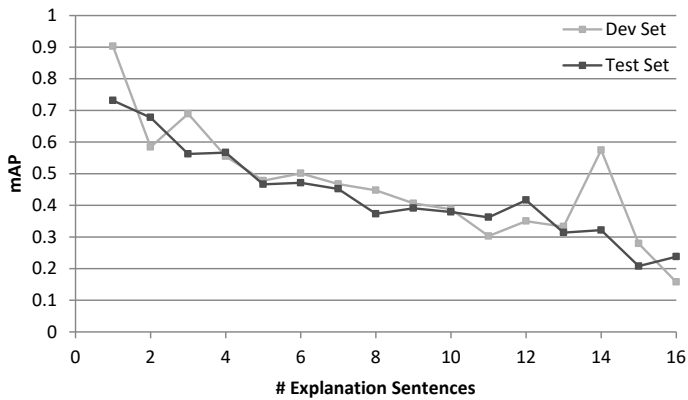

Figure 5: Percentage *mAP* of the SVM$^{reg}$ system on Development data (in light gray) and Test data (in dark gray), respectively, on different length explanation sentences.

## 8. Conclusion

In this work, we have investigated a knowledge-rich features-based approach for preference ordering of facts to explain the correct answer to elementary science questions. With the goal of creating meaningful unifications of $(q, ca, f)$ triples, we have investigated six different feature categories targeted to the domain at hand at varying lexical and semantic information representations. Further, our evaluations of regression versus learning-to-rank machine learning systems for preference ordering offers a new observation of the applicability of pointwise versus pairwise approaches (Melnikov et al. 2016; Fürnkranz and Hüllermeier 2010; Kamishima et al. 2010 2005).

Further, we have reported a detailed empirical analysis of our system on the task of explanation regeneration against nine existing reference systems. We have found that when provided with domain-targeted features, SVMs can outperform BERT-based neural approaches (Banerjee 2019). However, neural model variants applied in computationally complex task formulations (Das et al. 2019) can far surpass our system performance. Deep learning models generally report the best performances in NLP tasks (Manning 2015), with the limitation that their computationally complex task formulations are not practically suited. Nevertheless, they have long since consistently proven better than SVMs, w.r.t. both task performance and practicality in many cases. Our results obtained with SVMs then, in light of the general deep learning success phenomenon, seems particularly interesting - it offers a renewed empirical perspective on SVMs for the task defined in the WorldTree corpus that aligns better with symbolic problem formulations. While hand-crafting features for SVMs requires the linguistic insights of the practitioner to model the dataset well, they are avoided in light of recent performance boosts obtained from the black-box neural models. Our work sheds light on the fact that systems based on explicit feature modeling can still contend with neural approaches.

Overall, in this work, we have revisited a more traditional natural language engineering approach of leveraging linguistic features that are human-designed at multiple levels of the text data including syntax, semantics, and context. Given that our model, which can outperform the state-of-the-art BERT-based models, is fairly explainable, our paper offers insights for promising directions for further task improvements or for task engineering directions based on highly informative features.